\pdfoutput=1

\documentclass[11pt]{article}

\usepackage[preprint]{acl}

\usepackage{times}
\usepackage{latexsym}

\usepackage[T1]{fontenc}

\usepackage[utf8]{inputenc}

\usepackage{microtype}

\usepackage{inconsolata}

\usepackage{graphicx}

\usepackage{amsmath,amssymb,amsfonts}
\usepackage{algorithmic}
\usepackage{algorithm}
\usepackage{graphicx}
\usepackage{textcomp}
\usepackage{hyperref}
\usepackage{multirow}
\usepackage{makecell}
\usepackage{rotating}
\usepackage{mathtools}
\usepackage{xspace}
\usepackage{bm} 
\usepackage{url}
\usepackage{subcaption}
\def\argmin#1{\underset{#1}{\text{arg min\ }}}

\newcommand{\tref}[1]{Table~\ref{#1}}

\newcommand{\xmath}[1] {\ensuremath{#1}\xspace}
\newcommand{\blmath}[1] {\xmath{\bm{#1}}}

\newcommand{\Loss} {\xmath{\mathcal{L}}}

\newcommand{\Model} {\xmath{\mathcal{M}}}

\newcommand{\E}{\blmath{E}}

\newcommand{\I}{\blmath{I}}

\newcommand{\U}{\blmath{U}}
\newcommand{\V}{\blmath{V}}

\newcommand{\X}{\blmath{X}}

\newcommand{\y}{\blmath{y}}

\renewcommand{\Re}{\mathbb{R}}

\newcommand{\ba}{\blmath{\theta}}

\providecommand{\desa}[1]       {\begin{equation}%
    \begin{aligned}#1\end{aligned}\end{equation}} 

\title{LoPT: Low-Rank Prompt Tuning for Parameter Efficient Language Models}

\author{Shouchang Guo\thanks{Corresponding author.}, Sonam Damani, Keng-hao Chang \\ \\
        Microsoft AI \\[0.25em]
        \texttt{\{shouchangguo,sodamani,kenchan\}@microsoft.com}
        }

\begin{document}
\maketitle
\begin{abstract}
In prompt tuning, a prefix or suffix text is added to the prompt, and the embeddings (soft prompts) or token indices (hard prompts) of the prefix/suffix are optimized to gain more control over language models for specific tasks. This approach eliminates the need for hand-crafted prompt engineering or explicit model fine-tuning. Prompt tuning is significantly more parameter-efficient than model fine-tuning, as it involves optimizing partial inputs of language models to produce desired outputs.

In this work, we aim to further reduce the amount of trainable parameters required for a language model to perform well on specific tasks. We propose Low-rank Prompt Tuning (LoPT), a low-rank model for prompts that achieves efficient prompt optimization. The proposed method demonstrates similar outcomes to full parameter prompt tuning while reducing the number of trainable parameters by a factor of 5. It also provides promising results compared to the state-of-the-art methods that would require 10 to 20 times more parameters.
\end{abstract}

\section{Introduction}


With the success of large language models \cite{touvron2023llama,achiam2023gpt,jiang2023mistral}, it has become increasingly important for language models (LMs) to handle instructions effectively for customized agents and tasks. 
There are three essential categories of methods to adapt pre-trained language models to specific and customized needs: prompt engineering, model fine-tuning, and prompt tuning.

Prompt engineering \cite{brown2020language,sanh2021multitask,chung2024scaling} 
involves crafting handcrafted prompts and faces the challenge of getting LMs to consistently produce desired outputs with few-shot instructions. 
This effort may be difficult to generalize or extend to new tasks.
Model fine-tuning \cite{raffel2020exploring} can perform very well for task-specific needs but requires explicit fine-tuning of a significant number of model parameters, even with parameter-efficient fine-tuning (PEFT) approaches \cite{liu2022few, hu2021lora}.

Prompt tuning (PT) \cite{li2021prefix,lester2021power,wen2024hard,shi2022toward,shin2020autoprompt,khashabi2021prompt} is a promising method that lies between prompt engineering and model fine-tuning. Instead of handcrafting prompts, it optimizes a small number of prompt embeddings or indices with training data and has demonstrated capabilities comparable to those of model fine-tuning approaches \cite{asai2022attempt,shi2023dept,wang2023multitask}. 

We focus on soft prompt tuning, which operates by adding a prefix or suffix to the existing inputs and optimizing the embeddings of this prefix or suffix. The embeddings, or the soft prompt matrix, has dimensions $n \times d$, where $n$ is the ``tokens" length of soft prompts, and $d$ is the embedding size. The soft prompt length $n$ can be task specific to achieve desired outcomes. For example, more sophisticated tasks might benefit from longer soft prompts that allow for more parameters to be optimized.

In this work, we introduce a low-rank modeling approach for the soft prompt matrix, which effectively reduces the number of trainable parameters in prompt tuning without compromising performance. We find that soft prompt matrices are inherently low-rank due to their dimensionality, and we apply further dimensionality reduction through our proposed method. We demonstrate that the number of parameters required for tuning LMs to meet specific task requirements can be minimal. Additionally, the number of trainable parameters can be easily controlled by adjusting the rank of the soft prompt matrix.

Our approach distinguishes itself from existing methods by directly imposing low-rank constraints on the entire soft prompt to be trained. While recent work \cite{shi2023dept} also explores low-rank matrices for prompt tuning, it restricts low-rankness to the differences or updates of a frozen baseline prompt, similar to the LoRA technique used in model fine-tuning \cite{hu2021lora}, and is only applied to a portion of the overall soft prompt.

Our primary contributions are:
\begin{itemize}
\item We introduce Low-rank Prompt Tuning (LoPT) that significantly reduces the number of trainable parameters required in prompt tuning. 
\item We achieve a 5-fold reduction in trainable parameters while maintaining performance comparable to the full-parameter prompt tuning.
\item We demonstrate the efficacy of our method across 5 diverse datasets, showing substantial improvements in parameter efficiency compared to existing methods.
\end{itemize}

Our proposed parameter-efficient method would be particularly beneficial for computationally demanding prompt tuning needs in sophisticated tasks and large language models.










\section{Method}
\subsection{Problem statement}

In soft prompt tuning \citep{lester2021power}, 
we add a prefix or suffix to the original prompt and optimize the embeddings of this prefix or suffix as trainable parameters using supervised training data to achieve task-specific predictions.

Given a language model $\Model$ with frozen network parameters $\ba$ and embedding matrix $\E \in \Re^{V \times d}$, where $V$ is the vocabulary size, $d$ is the embedding size, with each row of $\E$ representing a token in the vocabulary. We optimize trainable embeddings $\X \in \Re^{n \times d}$ of the prefix, where $n$ is the number of soft tokens. The optimization problem can be formulated as:
\desa{
\begin{array}{ll}
\argmin{\X}
\sum_i \Loss\left(\Model\left(\left[\X;\I_i\right]; \ba \right),\y_i \right),
\end{array}
\label{eq1}
}
where $\Loss$ is the loss function for the task.
For the $i$-th training sample, $\I_i \in \Re^{t \times d}$ denotes tokenized embeddings of the original model input with sequence length $t$, and $\y_i$ is the label associated with this sample.

\subsection{Our Low-Rank Prompt Tuning (LoPT)}
Recent work \citep{lester2021power,shi2023dept} demonstrates that prompt tuning could yield performance comparable to parameter-efficient model fine-tuning methods \citep{hu2021lora} with a significantly smaller amount of learnable parameters. In this work, we push the boundaries by exploring parameter-efficient prompt tuning to further reduce the number of trainable parameters without compromising accuracy.

Because the prefix or suffix length $n$ is often significantly smaller that the embedding dimension $d$ in prompt tuning, the rank of the soft prompt matrix $\X$ would inherently be constrained by $n$, making $\X$ low-rank. The potential similarity between neighboring embeddings in a prompt could also suggest that $\X$ is low-rank. Therefore, we explore this potential and impose constraints on $\X$ for dimensionality reduction and more efficient prompt tuning. 

We propose two low-rank approximations for modeling $\X$. The proposed methods could drastically reduce the number of learnable parameters while maintaining performance comparable to full-parameter prompt tuning.





\label{fig1} 

\subsubsection{LoPT-1}
For effective prompt tuning with a reduced and adjustable number of parameters, we propose to decomposite the low-rank prompt matrix $\X \in \Re^{n \times d}$ as:
\begin{equation}
\X = \U\V.
\end{equation}
In this formulation, $\U \in \Re^{n \times r}$ and $\V \in \Re^{r \times d}$ are the new trainable matrices.
We train $\U$ and $\V$ simultaneously, transforming the prompt tuning optimization problem to the following:
\desa{
\begin{array}{ll}
\argmin{\U,\,\V}
\sum_i \Loss\left(\Model\left(\left[\U\V;\I_i\right]; \ba \right),\y_i \right).
\end{array}
}
We initialize both $\U$ and $\V$ with uniform random values in the range of [-0.5, 0.5] at the beginning of training.

The number of trainable parameters is reduced to $r(n + d)$. As $n \ll d$, the total number of parameters can be significantly reduced compared to the original $nd$, especially with adjustable choices of $r < n$.

\subsubsection{LoPT-2}
we also introduce an empirical mapping scheme for the low-rank approximation of $\X$,
employing learnable linear projections and nonlinear thresholding operation to achieve effects analogous to singular value thresholding \citep{cai2010singular} and with reduced number of parameters for optimization. Specifically, we construct $\X$ as:
\begin{equation}
\X = \sigma(\X_0 \U)\V,
\end{equation}
where $\X_0 \in \Re^{n \times d}$ is a random initialization of $\X$, $\U \in \Re^{d \times r}$ and $V \in \Re^{r \times d}$ are linear projection matrices. $\sigma(\cdot) = \mathrm{max}(\cdot, 0)$ represents the nonlinear thresholding operation that filters out negative values. 
Similar to LoPT-1, $\U$ and $\V$ are randomly initialized and optimized with function
\desa{
\begin{array}{ll}
\argmin{\U,\,\V}
\sum_i \Loss\left(\Model\left(\left[\sigma(\X_0 \U)\V;\I_i\right]; \ba \right),\y_i \right).
\end{array}
}

The number of trainable parameters becomes $2rd$ rather than $nd$. By choosing a smaller projected dimension $r < n/2$, we can easily reduce redundancy in trainable parameters and improve time and memory efficiency. 
It is worth noting that for $n \ll d$, LoPT-1 is more parameter efficient than LoPT-2. 

\textbf{Implementation Simplification}
The proposed LoPT-2 mapping for $\X$ improves parameter efficiency, and we propose a straightforward implementation. We use two linear layers for the linear projections $U$ and $V$, and apply an ELU \citep{clevert2015fast} function for the nonlinear thresholding operator $\sigma(\cdot)$. Empirically, we found that ELU performs better than ReLU \citep{nair2010rectified,fukushima1969visual} and GELU \citep{hendrycks2016gaussian}.

We demonstrate that the proposed low-rank modeling and formulations yield effective parameter reduction with promising outcomes.






\begin{table}
  \begin{tabular}{lc|cc}
    \hline
   \textbf{Method} & \textbf{\# Params} & \textbf{SST-2} & \textbf{AGNews} \\
    \hline
  No LoPT & 12.8k & 92.8 & 91.8 \\
  LoPT-1 (ours) & 2.58k & 92.1 & 91.9 \\
  LoPT-2 (ours) & 5.12k & 90.9 & 90.0 \\
  \hline
\end{tabular}
\caption{\label{tab1}
Accuracy (\%) on the SST-2 and AGNews validation sets compares the proposed LoPT-1 and LoPT-2 to the baseline soft prompt tuning without low-rank factorization (No LoPT). The language model used is GPT-2 large with embedding dimension $d = 1280$, and prompt length $n = 10$. We set the rank $r = 2$ for both LoPT-1 and LoPT-2, and calculate the \# of parameters accordingly.}
\end{table}

\section{Experiments}
\subsection{Experiment Setup}

\begin{table*}
  \centering
  \begin{tabular}{lc|ccccc}
    \hline
    \textbf{Method} & \textbf{\# Params} & \textbf{SST-2} & \textbf{BoolQ} & \textbf{RTE} & \textbf{WiC} & \textbf{CB}\\
    \hline
    Fine-tuning\textsuperscript{1} & 220M & 94.6 & 81.1 & 71.9 & 70.2 & 85.7 \\
    LoRA\textsuperscript{2} & 3.8M & 94.3 & 81.3 & 75.5 & 68.3 & 92.9 \\
    PT\textsuperscript{3} & 76.8k & 91.9 & 63.7 & 78.8 & 50.8 & 67.9 \\
    DePT\textsuperscript{3} & 76.8k & 94.2 & 79.3 & 79.1 & 68.7 & 92.9 \\
    LoPT-1 (ours) & 3.94k & 92.9 & 76.5 & 73.8 & 55.1 & 90.4 \\
    LoPT-2 (ours) & 7.68k & 92.4 & 75.5 & 74.3 & 62.7 & 74.0 \\
    \hline
  \end{tabular}
  \caption{\label{tab2}
  Accuracy (\%) on the SST-2 and SuperGLUE benchmarks for classification tasks. The language model is T5-Base with embedding dimension $d = 768$. We set the rank $r = 5$ and soft prompt length $n = 20$ for both LoPT-1 and LoPT-2.
  Comparisons including Fine-tuning\textsuperscript{1} from \citep{asai2022attempt}, LoRA\textsuperscript{2} from \citep{sung2022lst}, PT\textsuperscript{3} and DePT\textsuperscript{3} are from \citep{shi2023dept}.
  }
\end{table*}

\begin{table}
  \begin{tabular}{llcc}
    \hline
  \textbf{Length} & \textbf{Rank} & \textbf{$\Delta$ \# Params} & \textbf{SST-2}\\
    \hline
    {$n=10$} & No LoPT & - & 92.8 \\
    \hline
 \multirow{3}{*}{$n=10$} & $r=1$ & -89.92\% & 90.5 \\
 & $r=2$ & -79.84\% & 92.1  \\
 & $r=5$ & -49.61\% & 92.1  \\
  \hline
 \multirow{3}{*}{$n=20$} & $r=1$ & -89.84\% & 91.4 \\
 & $r=2$ & -79.69\% & 92.8  \\
 & $r=5$ & -49.22\% & \textbf{92.9}  \\
  \hline
\multirow{3}{*}{$n=30$} & $r=1$ & -89.77\% & 90.9 \\
 & $r=2$ & -79.53\% & 92.2  \\
 & $r=5$ & -48.83\% & 92.1  \\
  \hline
\end{tabular}
\caption{\label{tab3}
Ablation study on LoPT-1: We evaluated various combinations of prompt length $n$ and rank $r$ using the SST-2 dataset and the GPT-2 large model. The numbers of trainable parameters are compared to the baseline prompt tuning, which has a fixed $n = 10$ and no low-rank approximations. The parameter reduction rate is represented by $\Delta$ \# Params. LoPT-1 with $n = 20$ and $r = 5$ achieves the highest accuracy (\%).}
\end{table}

\textbf{Datasets}
We evaluate the proposed method on classification tasks using various datasets in English: the sentiment analysis task SST-2 \citep{socher-etal-2013-recursive}, the 4-way topic classification task AGNews \citep{zhang2015character}, and datasets in the SuperGLUE benchmark \citep{wang2019superglue}. These include BoolQ \citep{clark2019boolq}, RTE \citep{giampiccolo2007third}, WiC \citep{pilehvar2018wic}, and CB \citep{de2019commitmentbank}.

\noindent \textbf{Training Details} 
The proposed low-rank factorizations, LoPT-1 and LoPT-2, are optimized using GPT-2 large (774M parameters, $d = 1280$) \citep{radford2019language} and T5-base (220M parameters, $d = 768$) \citep{raffel2020exploring} models. We build upon the settings in \citep{ding2021openprompt,wen2024hard}, and optimize the prompts using the Adafactor optimizer \citep{shazeer2018adafactor} with a learning rate of 0.3. We apply soft prompt length $n$ of 10 or 20, and batch size of 8 for SuperGLUE datasets, and 16 for other data. 

We set the rank parameter $r$ of LoPT-1 or LoPT-2 to $\lfloor \frac{n}{4} \rfloor$ for most experiments to achieve the desired level of trainable parameter reduction. In the case of prompt tuning without our proposed low-rank approximations, the number of trainable parameters is $nd$. For LoPT-1, the number of learnable parameters is $r(n+d)$. For LoPT-2, the trainable parameter amount is $2dr$.

\subsection{Comparisons and Results}
We compare the proposed parameter efficient approaches to vanilla soft prompt tuning using the GPT-2 large model, and evaluate their effectiveness with SST-2 and AGNews datasets. As presented in \tref{tab1}, LoPT-1 significantly reduces the number of trainable parameters from 12.8k to 2.58k, while maintaining accuracy levels comparable to full parameter prompt tuning. LoPT-2 achieves a 60\% reduction in parameters and successfully preserves classification accuracy for both binary and multi-class classification tasks.

Our methods are compared against a variety of baselines including Fine-tuning, LoRA \citep{hu2021lora}, PT \citep{lester2021power}, and DePT \citep{shi2023dept} using the T5-base model. As shown in \tref{tab2}, LoPT-1 and LoPT-2 demonstrate promising performance, achieving reductions in trainable parameters by factors of 20 and 10, respectively.  This marks a significant efficiency improvement over existing prompt tuning approaches, which are already noted for their high parameter efficiency.

It is noteworthy that LoPT-1 outperforms LoPT-2 on the CB dataset, while LoPT-2 excels over LoPT-1 on the WiC dataset. This suggests that both approaches could be strategically exploited to tailor the desired low-rank formation for optimal performance on specific tasks. 


\subsection{Ablation Study}
Using the SST-2 task and the GPT-2 large model, \tref{tab3} presents the accuracy of LoPT-1 with varying prompt lengths $n$ and ranks $r$ for the low-rank factorization. We observe that an increased prompt length does not necessarily lead to improved outcomes, and the combination of $n = 20$ with $r = 5$ or $r = 2$ yield the highest accuracy. Given that $n$ is much smaller than $d$, the number of trainable parameters is primarily controlled by the rank parameter $r$ in LoPT, which can be easily adjusted to achieve parameter reduction.

\subsection{Limitations}
This work relies on the low-rank hypothesis and may not be effective when the prompt matrix is not low-rank. Regarding the performance of the proposed methods, further improvements could be achieved through hyper-parameter tuning.



\section{Conclusion}
In this work, we propose Low-rank Prompt Tuning (LoPT), a low-rank formulation of prompts that significantly reduces the number of trainable parameters for parameter-efficient prompt tuning of language models. We demonstrate that LoPT can decrease the number of trainable parameters by a factor of 10 or 20 while achieving promising performance across various datasets.

The proposed parameter-efficient method could be particularly beneficial for sophisticated tasks and large language models, where longer soft prompts are increasingly important for effective prompt tuning.

\bibliography{6ref}

\begin{thebibliography}{34}
\providecommand{\natexlab}[1]{#1}

\bibitem[{Achiam et~al.(2023)Achiam, Adler, Agarwal, Ahmad, Akkaya, Aleman, Almeida, Altenschmidt, Altman, Anadkat et~al.}]{achiam2023gpt}
Josh Achiam, Steven Adler, Sandhini Agarwal, Lama Ahmad, Ilge Akkaya, Florencia~Leoni Aleman, Diogo Almeida, Janko Altenschmidt, Sam Altman, Shyamal Anadkat, et~al. 2023.
\newblock Gpt-4 technical report.
\newblock \emph{arXiv preprint arXiv:2303.08774}.

\bibitem[{Asai et~al.(2022)Asai, Salehi, Peters, and Hajishirzi}]{asai2022attempt}
Akari Asai, Mohammadreza Salehi, Matthew~E Peters, and Hannaneh Hajishirzi. 2022.
\newblock Attempt: Parameter-efficient multi-task tuning via attentional mixtures of soft prompts.
\newblock \emph{arXiv preprint arXiv:2205.11961}.

\bibitem[{Brown et~al.(2020)Brown, Mann, Ryder, Subbiah, Kaplan, Dhariwal, Neelakantan, Shyam, Sastry, Askell et~al.}]{brown2020language}
Tom Brown, Benjamin Mann, Nick Ryder, Melanie Subbiah, Jared~D Kaplan, Prafulla Dhariwal, Arvind Neelakantan, Pranav Shyam, Girish Sastry, Amanda Askell, et~al. 2020.
\newblock Language models are few-shot learners.
\newblock \emph{Advances in neural information processing systems}, 33:1877--1901.

\bibitem[{Cai et~al.(2010)Cai, Cand{\`e}s, and Shen}]{cai2010singular}
Jian-Feng Cai, Emmanuel~J Cand{\`e}s, and Zuowei Shen. 2010.
\newblock A singular value thresholding algorithm for matrix completion.
\newblock \emph{SIAM Journal on optimization}, 20(4):1956--1982.

\bibitem[{Chung et~al.(2024)Chung, Hou, Longpre, Zoph, Tay, Fedus, Li, Wang, Dehghani, Brahma et~al.}]{chung2024scaling}
Hyung~Won Chung, Le~Hou, Shayne Longpre, Barret Zoph, Yi~Tay, William Fedus, Yunxuan Li, Xuezhi Wang, Mostafa Dehghani, Siddhartha Brahma, et~al. 2024.
\newblock Scaling instruction-finetuned language models.
\newblock \emph{Journal of Machine Learning Research}, 25(70):1--53.

\bibitem[{Clark et~al.(2019)Clark, Lee, Chang, Kwiatkowski, Collins, and Toutanova}]{clark2019boolq}
Christopher Clark, Kenton Lee, Ming-Wei Chang, Tom Kwiatkowski, Michael Collins, and Kristina Toutanova. 2019.
\newblock Boolq: Exploring the surprising difficulty of natural yes/no questions.
\newblock \emph{arXiv preprint arXiv:1905.10044}.

\bibitem[{Clevert et~al.(2015)Clevert, Unterthiner, and Hochreiter}]{clevert2015fast}
Djork-Arn{\'e} Clevert, Thomas Unterthiner, and Sepp Hochreiter. 2015.
\newblock Fast and accurate deep network learning by exponential linear units (elus).
\newblock \emph{arXiv preprint arXiv:1511.07289}.

\bibitem[{De~Marneffe et~al.(2019)De~Marneffe, Simons, and Tonhauser}]{de2019commitmentbank}
Marie-Catherine De~Marneffe, Mandy Simons, and Judith Tonhauser. 2019.
\newblock The commitmentbank: Investigating projection in naturally occurring discourse.
\newblock In \emph{proceedings of Sinn und Bedeutung}, volume~23, pages 107--124.

\bibitem[{Ding et~al.(2021)Ding, Hu, Zhao, Chen, Liu, Zheng, and Sun}]{ding2021openprompt}
Ning Ding, Shengding Hu, Weilin Zhao, Yulin Chen, Zhiyuan Liu, Hai-Tao Zheng, and Maosong Sun. 2021.
\newblock Openprompt: An open-source framework for prompt-learning.
\newblock \emph{arXiv preprint arXiv:2111.01998}.

\bibitem[{Fukushima(1969)}]{fukushima1969visual}
Kunihiko Fukushima. 1969.
\newblock Visual feature extraction by a multilayered network of analog threshold elements.
\newblock \emph{IEEE Transactions on Systems Science and Cybernetics}, 5(4):322--333.

\bibitem[{Giampiccolo et~al.(2007)Giampiccolo, Magnini, Dagan, and Dolan}]{giampiccolo2007third}
Danilo Giampiccolo, Bernardo Magnini, Ido Dagan, and William~B Dolan. 2007.
\newblock The third pascal recognizing textual entailment challenge.
\newblock In \emph{Proceedings of the ACL-PASCAL workshop on textual entailment and paraphrasing}, pages 1--9.

\bibitem[{Hendrycks and Gimpel(2016)}]{hendrycks2016gaussian}
Dan Hendrycks and Kevin Gimpel. 2016.
\newblock Gaussian error linear units (gelus).
\newblock \emph{arXiv preprint arXiv:1606.08415}.

\bibitem[{Hu et~al.(2021)Hu, Shen, Wallis, Allen-Zhu, Li, Wang, Wang, and Chen}]{hu2021lora}
Edward~J Hu, Yelong Shen, Phillip Wallis, Zeyuan Allen-Zhu, Yuanzhi Li, Shean Wang, Lu~Wang, and Weizhu Chen. 2021.
\newblock Lora: Low-rank adaptation of large language models.
\newblock \emph{arXiv preprint arXiv:2106.09685}.

\bibitem[{Jiang et~al.(2023)Jiang, Sablayrolles, Mensch, Bamford, Chaplot, Casas, Bressand, Lengyel, Lample, Saulnier et~al.}]{jiang2023mistral}
Albert~Q Jiang, Alexandre Sablayrolles, Arthur Mensch, Chris Bamford, Devendra~Singh Chaplot, Diego de~las Casas, Florian Bressand, Gianna Lengyel, Guillaume Lample, Lucile Saulnier, et~al. 2023.
\newblock Mistral 7b.
\newblock \emph{arXiv preprint arXiv:2310.06825}.

\bibitem[{Khashabi et~al.(2021)Khashabi, Lyu, Min, Qin, Richardson, Welleck, Hajishirzi, Khot, Sabharwal, Singh et~al.}]{khashabi2021prompt}
Daniel Khashabi, Shane Lyu, Sewon Min, Lianhui Qin, Kyle Richardson, Sean Welleck, Hannaneh Hajishirzi, Tushar Khot, Ashish Sabharwal, Sameer Singh, et~al. 2021.
\newblock Prompt waywardness: The curious case of discretized interpretation of continuous prompts.
\newblock \emph{arXiv preprint arXiv:2112.08348}.

\bibitem[{Lester et~al.(2021)Lester, Al-Rfou, and Constant}]{lester2021power}
Brian Lester, Rami Al-Rfou, and Noah Constant. 2021.
\newblock The power of scale for parameter-efficient prompt tuning.
\newblock \emph{arXiv preprint arXiv:2104.08691}.

\bibitem[{Li and Liang(2021)}]{li2021prefix}
Xiang~Lisa Li and Percy Liang. 2021.
\newblock Prefix-tuning: Optimizing continuous prompts for generation.
\newblock \emph{arXiv preprint arXiv:2101.00190}.

\bibitem[{Liu et~al.(2022)Liu, Tam, Muqeeth, Mohta, Huang, Bansal, and Raffel}]{liu2022few}
Haokun Liu, Derek Tam, Mohammed Muqeeth, Jay Mohta, Tenghao Huang, Mohit Bansal, and Colin~A Raffel. 2022.
\newblock Few-shot parameter-efficient fine-tuning is better and cheaper than in-context learning.
\newblock \emph{Advances in Neural Information Processing Systems}, 35:1950--1965.

\bibitem[{Nair and Hinton(2010)}]{nair2010rectified}
Vinod Nair and Geoffrey~E Hinton. 2010.
\newblock Rectified linear units improve restricted boltzmann machines.
\newblock In \emph{Proceedings of the 27th international conference on machine learning (ICML-10)}, pages 807--814.

\bibitem[{Pilehvar and Camacho-Collados(2018)}]{pilehvar2018wic}
Mohammad~Taher Pilehvar and Jose Camacho-Collados. 2018.
\newblock Wic: the word-in-context dataset for evaluating context-sensitive meaning representations.
\newblock \emph{arXiv preprint arXiv:1808.09121}.

\bibitem[{Radford et~al.(2019)Radford, Wu, Child, Luan, Amodei, Sutskever et~al.}]{radford2019language}
Alec Radford, Jeffrey Wu, Rewon Child, David Luan, Dario Amodei, Ilya Sutskever, et~al. 2019.
\newblock Language models are unsupervised multitask learners.
\newblock \emph{OpenAI blog}, 1(8):9.

\bibitem[{Raffel et~al.(2020)Raffel, Shazeer, Roberts, Lee, Narang, Matena, Zhou, Li, and Liu}]{raffel2020exploring}
Colin Raffel, Noam Shazeer, Adam Roberts, Katherine Lee, Sharan Narang, Michael Matena, Yanqi Zhou, Wei Li, and Peter~J Liu. 2020.
\newblock Exploring the limits of transfer learning with a unified text-to-text transformer.
\newblock \emph{Journal of machine learning research}, 21(140):1--67.

\bibitem[{Sanh et~al.(2021)Sanh, Webson, Raffel, Bach, Sutawika, Alyafeai, Chaffin, Stiegler, Scao, Raja et~al.}]{sanh2021multitask}
Victor Sanh, Albert Webson, Colin Raffel, Stephen~H Bach, Lintang Sutawika, Zaid Alyafeai, Antoine Chaffin, Arnaud Stiegler, Teven~Le Scao, Arun Raja, et~al. 2021.
\newblock Multitask prompted training enables zero-shot task generalization.
\newblock \emph{arXiv preprint arXiv:2110.08207}.

\bibitem[{Shazeer and Stern(2018)}]{shazeer2018adafactor}
Noam Shazeer and Mitchell Stern. 2018.
\newblock Adafactor: Adaptive learning rates with sublinear memory cost.
\newblock In \emph{International Conference on Machine Learning}, pages 4596--4604. PMLR.

\bibitem[{Shi et~al.(2022)Shi, Han, Gonen, Holtzman, Tsvetkov, and Zettlemoyer}]{shi2022toward}
Weijia Shi, Xiaochuang Han, Hila Gonen, Ari Holtzman, Yulia Tsvetkov, and Luke Zettlemoyer. 2022.
\newblock Toward human readable prompt tuning: Kubrick's the shining is a good movie, and a good prompt too?
\newblock \emph{arXiv preprint arXiv:2212.10539}.

\bibitem[{Shi and Lipani(2023)}]{shi2023dept}
Zhengxiang Shi and Aldo Lipani. 2023.
\newblock Dept: Decomposed prompt tuning for parameter-efficient fine-tuning.
\newblock \emph{arXiv preprint arXiv:2309.05173}.

\bibitem[{Shin et~al.(2020)Shin, Razeghi, Logan~IV, Wallace, and Singh}]{shin2020autoprompt}
Taylor Shin, Yasaman Razeghi, Robert~L Logan~IV, Eric Wallace, and Sameer Singh. 2020.
\newblock Autoprompt: Eliciting knowledge from language models with automatically generated prompts.
\newblock \emph{arXiv preprint arXiv:2010.15980}.

\bibitem[{Socher et~al.(2013)Socher, Perelygin, Wu, Chuang, Manning, Ng, and Potts}]{socher-etal-2013-recursive}
Richard Socher, Alex Perelygin, Jean Wu, Jason Chuang, Christopher~D. Manning, Andrew Ng, and Christopher Potts. 2013.
\newblock \href {D13-1170} {Recursive deep models for semantic compositionality over a sentiment treebank}.
\newblock pages 1631--1642, Seattle, Washington, USA.

\bibitem[{Sung et~al.(2022)Sung, Cho, and Bansal}]{sung2022lst}
Yi-Lin Sung, Jaemin Cho, and Mohit Bansal. 2022.
\newblock Lst: Ladder side-tuning for parameter and memory efficient transfer learning.
\newblock \emph{Advances in Neural Information Processing Systems}, 35:12991--13005.

\bibitem[{Touvron et~al.(2023)Touvron, Martin, Stone, Albert, Almahairi, Babaei, Bashlykov, Batra, Bhargava, Bhosale et~al.}]{touvron2023llama}
Hugo Touvron, Louis Martin, Kevin Stone, Peter Albert, Amjad Almahairi, Yasmine Babaei, Nikolay Bashlykov, Soumya Batra, Prajjwal Bhargava, Shruti Bhosale, et~al. 2023.
\newblock Llama 2: Open foundation and fine-tuned chat models.
\newblock \emph{arXiv preprint arXiv:2307.09288}.

\bibitem[{Wang et~al.(2019)Wang, Pruksachatkun, Nangia, Singh, Michael, Hill, Levy, and Bowman}]{wang2019superglue}
Alex Wang, Yada Pruksachatkun, Nikita Nangia, Amanpreet Singh, Julian Michael, Felix Hill, Omer Levy, and Samuel Bowman. 2019.
\newblock Superglue: A stickier benchmark for general-purpose language understanding systems.
\newblock \emph{Advances in neural information processing systems}, 32.

\bibitem[{Wang et~al.(2023)Wang, Panda, Karlinsky, Feris, Sun, and Kim}]{wang2023multitask}
Zhen Wang, Rameswar Panda, Leonid Karlinsky, Rogerio Feris, Huan Sun, and Yoon Kim. 2023.
\newblock Multitask prompt tuning enables parameter-efficient transfer learning.
\newblock \emph{arXiv preprint arXiv:2303.02861}.

\bibitem[{Wen et~al.(2024)Wen, Jain, Kirchenbauer, Goldblum, Geiping, and Goldstein}]{wen2024hard}
Yuxin Wen, Neel Jain, John Kirchenbauer, Micah Goldblum, Jonas Geiping, and Tom Goldstein. 2024.
\newblock Hard prompts made easy: Gradient-based discrete optimization for prompt tuning and discovery.
\newblock \emph{Advances in Neural Information Processing Systems}, 36.

\bibitem[{Zhang et~al.(2015)Zhang, Zhao, and LeCun}]{zhang2015character}
Xiang Zhang, Junbo Zhao, and Yann LeCun. 2015.
\newblock Character-level convolutional networks for text classification.
\newblock \emph{Advances in neural information processing systems}, 28.

\end{thebibliography}

\end{document}